\documentclass[smallextended]{paper}       % onecolumn (second format)
\usepackage[utf8]{inputenc}
\usepackage{authblk}
\usepackage{hyperref}
\usepackage{breakurl}
\usepackage{mathptmx}       % selects Times Roman as basic font
\usepackage{helvet}         % selects Helvetica as sans-serif font
\usepackage{courier}        % selects Courier as typewriter font
\usepackage{type1cm}        % activate if the above 3 fonts are
\usepackage{makeidx}         % allows index generation
\usepackage{graphicx}        % standard LaTeX graphics tool
\usepackage{multicol}        % used for the two-column index
\usepackage[bottom]{footmisc}% places footnotes at page bottom

\usepackage{marginnote,soul,cancel}

\usepackage{amsfonts,amssymb,amsmath}
\usepackage{epsfig,rotating,framed,enumerate,listings,subfigure}
\usepackage{nicefrac}
\usepackage{wrapfig}

\usepackage{algorithm}
\usepackage[noend]{algpseudocode}
\newcommand{\Bildeinbinden}[2] {\setlength{\epsfxsize}{#1}\epsfbox{#2}}

\allowdisplaybreaks
\begin{document}

\title{ReLU activated Multilayer Neural Networks trained with Mixed
Integer Linear Programs}

\author{Steffen Goebbels\footnote{Niederrhein University of Applied Sciences, Faculty of Electrical Engineering
and Computer Science, Institute for Pattern
Recognition, D-47805 Krefeld, Germany, steffen.goebbels@hsnr.de}}

\maketitle
\begin{center}
The third uploaded version is a preprint of paper with title ``Training of ReLU
Activated Multilayer Neural Networks with Mixed Integer Linear Programs'' that
has been published by the Faculty of Electrical Engineering and Computer Science of the
Niederrhein University of Applied Sciences:\\ 
\url{https://www.hs-niederrhein.de/elektrotechnik-informatik/technische-berichte/}
\end{center}

\begin{abstract}
In this paper, it is demonstrated through a case study that multilayer
feedforward neural networks activated by ReLU functions can in principle be trained iteratively with 
Mixed Integer Linear Programs (MILPs) as follows.
Weights are determined with batch
learning.
Multiple iterations are used per batch of training data. 
In each iteration, the algorithm starts at the output layer and propagates 
information back to the first hidden layer to adjust the weights using MILPs or 
Linear Programs. For each layer, the goal is to minimize the difference between
its output and the corresponding target output. The target output of the last
(output) layer is equal to the ground truth. The target output of a previous layer is defined 
as the adjusted input of the following layer. For a given layer, weights are
computed by solving a MILP. Then, except for the first
hidden layer, the input values are also
modified with a MILP to better
match the layer outputs to their corresponding target outputs.
The method was tested and compared with Tensorflow/Keras (Adam optimizer) 
using two simple networks on the MNIST dataset containing handwritten digits.
Accuracies of the same magnitude as with Tensorflow/Keras were achieved.

%\subclass{41A25 \and 41A50 \and 62M45}
\end{abstract}

\section{Introduction}
\allowdisplaybreaks

Neural networks typically learn by adjusting weights using nonlinear
optimization in a training phase. Variants of gradient
descent are often used. These techniques require ``some''
differentiability of the error functional.
Therefore, piecewise linear activation functions like  
the Rectified Linear Unit (ReLU)
$$\sigma(x):=\max\{0,x\}$$
or
the Heaviside function, that are not differentiable at the origin, raise the
question of whether linear and mixed integer linear programming techniques are also suitable for network training.

%07.10. umformuliert
Learning to near
optimality can be done with Linear Programs (LP) of exponential size
for certain network architectures, see \cite{Bienstock18}. But this is not
applicable in practice.
Mixed Integer Linear Programs (MILPs) are
proposed in \cite{Fischetti18} to find inputs of ReLU networks that
maximize unit activation. This can help to understand the features
computed in the network. To this end, the weights are not variable. The output
of a neuron is modeled by the same constraints as in
\cite{Serra17}, where MILPs are used to count maximum numbers of linear regions
in outputs of ReLU networks. In order to find vulnerabilities, a trained binary
neural network is attacked by a MILP in \cite{Khalil18}. This MILP computes
inputs for which the network fails to predict. In both this MILP and in \cite{Anderson20}, the
weights are also considered as constants. Another approach to evaluate the robustness of networks is described in
\cite{tjeng2017evaluating}. A network layout consisting of nodes and edges is
optimized with a MILP in \cite{Dua10}.

The present work investigates the suitability of training with MILPs,
i.e.~in contrast to the previously mentioned works, network weights are now
variables of the optimization problem.
Oracle Inc.~holds US patent \cite{Patent}, which protects the
idea of using MILPs for training parts of (deep) neural networks. The solution
described in this patent works in a scenario with piecewise constant
activation functions for hidden neurons as well as piecewise linear activation
functions on the output layer. Without additional algorithmic intervention, it
does not work when values of weights (that are variables in the optimization
problem) have to be multiplied.
This is the case when activation functions that are not piecewise constant are used on successive layers. 
Thus, additional considerations are required for ReLU-activated
networks to use linear optimization methods.

\begin{figure*}[htb]
\begin{minipage}{0.55\columnwidth}
\begin{center}
      \Bildeinbinden{0.55\columnwidth}{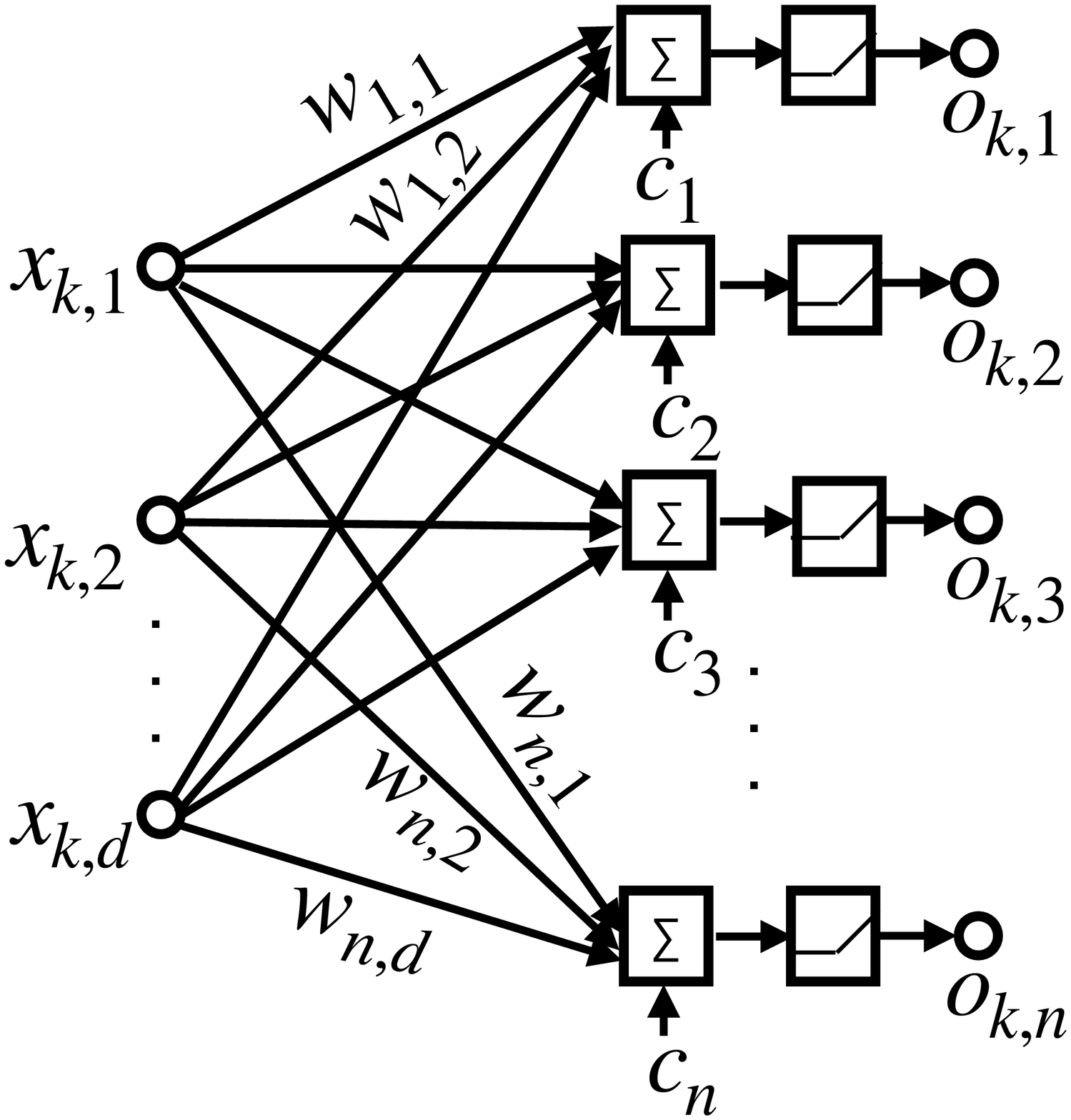}
\end{center}  
\end{minipage}
\begin{minipage}{0.43\columnwidth}
By applying the ReLU function $\sigma$
to each component of the vector
$W\vec{x}+\vec{c}$, the output $\vec{o}$ is obtained ($\vec{x}\in\mathbb{R}^d$,
$\vec{o}\in\mathbb{R}^n$, $\vec{c}\in\mathbb{R}^n$, $W\in\mathbb{R}^{n\times d}$):
$$\vec{o} = \sigma(W\vec{x}+\vec{c}).$$
\end{minipage}   
\caption{Building block of ReLU-activated feedforward
network: blocks can be concatenated
to realize a deep network. The output $\vec{o}$ of
a layer then becomes the input $\vec{x}$ of the next
layer, i.e., the dimensions of subsequent building blocks have
to fit. The building blocks do not share weights.\label{figureblock}}
\end{figure*}

\begin{algorithm*}[htb]
\caption{Iterative backpropagation-like learning with MILPs\label{algMILP}}
{\small
\begin{algorithmic}
\Procedure {learn weights}{training\_input\_data, training\_ground\_truth\_data}
\State Randomly initialize all weights
\State accuracy := 0, last\_accuracy := -1,
target\_values := training\_ground\_truth\_data
\While{accuracy $>$ last\_accuracy}
  \State last\_accuracy := accuracy
  \State Compute all neuron outputs $\vec{o}$ for training\_input\_data
  \For{i $:=$ number of output layer {\bf back to} number 1 of first hidden
  layer} \State Update weights of (output or hidden) layer i with
      LP/MILP, see Section \ref{secMILP1}:
     \State\quad  Minimize $l^1$ norm of differences
     between output values of layer i and% for all instances of
     %training\_input\_data.
     \State\quad target\_values. Input values of layer i are fixed, weights are
     variables.
     \If{i $>$ 1}
           \State Compute optimal input values $\vec{x}$ of this layer (which
           are output values $\vec{o}$ of the 
           \State preceding layer) using a
                second LP/MILP, see Section
                \ref{secMILP2}: %for all instances of training data set
            \State\quad Minimize $l^1$ norm of differences
            between output values of layer i
            \State\quad and target\_values.
            Weights of the layer are now fixed. 
            \State\quad Input values are variables.
         \State target\_values := computed optimal input values
     \EndIf
  \EndFor
  \State For updated weights and training\_input\_data, update inputs and
  outputs of all neurons 
  \State Re-compute accuracy 
\EndWhile
\If{accuracy $<$ 1}
\State Update weights with those belonging to best accuracy that occurred in
while-loop
\State Finally optimize weights of the last layer, see LP in
Section \ref{seclastlayer}.
\EndIf
\EndProcedure
\end{algorithmic}}
\end{algorithm*}
We investigate a backpropagation-like algorithm (see Algorithm \ref{algMILP}) to
iteratively train a ReLU network with LPs and MILPs.
The prerequisite is a neural network with ReLU activation that
is a concatenation of building blocks, as shown in Figure \ref{figureblock}. 
All hidden layers and the output layer consist of such a building block. 
Edges can be removed by setting their weights fixed to zero. In addition, the
equality of weights can be specified. This makes it possible, for example, to realize convolutional layers. 
The input layer only passes
values to the building block of the first hidden layer.
Most deep neural networks follow this architecture but use a
different activation function like softmax on the output layer. For simplicity,
we also use ReLU there.

To evaluate the algorithm, we select the MNIST
dataset\footnote{http://yann.lecun.com/exdb/mnist/}, see \cite{LeCun98}, that
consists of $60.000$ images ($28\times 28$ pixels) of handwritten
digits for training and $10.000$ digits for testing, in connection with two
small example instances of the discussed network.
One instance consists of 784-8-8-8-10 neurons on five layers (three
hidden fully connected layers), cf.~\cite[DNN1]{Fischetti18}. The values of
ten output neurons encode the recognized number. Another example is a
49-25-10 network with a convolutional (single feature-map) and a subsequent
fully connected layer.
To apply the network, the images are downsampled to a size of $7\times 7$ gray
values by taking the mean values of $4\times 4$ regions. The size of the
convolution kernel is $3\times 3$, all offsets $c_j$ are set to zero. For both
network instances, the index of an output neuron whose output is closest to one
represents the detected number.\label{defacc}
%If a output value exceeds 0.5 then the index of the neuron represents this
% number. We regard the output as a wrong detection result if more or less than
% one value is above 0.5 or if the detected number is different to the ground truth. 
The accuracy is
the relative number of true detections.

On batches (subsets) of training data (which sadly must be small
because of runtimes), we determine the weights using Algorithm \ref{algMILP}, that 
consists of three MILPs.
%
%the inputs from the training
%dataset.
%To be more specific:
%Each input of the training dataset is seen as equidistant sampled values on
%% $[0, 1]$ that define a piecewise linear continuous interpolation function. In
% turn, we
%equidistantly sample this functions to obtain initial neuron outputs of each
%layer. Note that the number of neurons on each layer might be different. The
%outputs of the input neurons correspond with the training input.
%The outer loop is not necessary, if there is only one hidden layer. Then
% weights are optimally chosen with the first MILP.
%
They are specified in the next section. Then the results are
compared with those of gradient descent as implemented by the widely
used Adam optimizer \cite{Adam}.

%bis hier

\section{Mixed Integer Linear Programs and Linear Programs}
\subsection{Computation of weights}\label{secMILP1}
We determine weights 
$W\in\mathbb{R}^{n\times d}$,
$$W=[w_{l,j}]_{l\in[n],\, j\in[d]:=\{1,\dots,d\}},$$ and
$\vec{c}\in\mathbb{R}^{n}$ (with components $c_j$) of one building block (see Figure \ref{figureblock}) with $d$
inputs and $n$ outputs.
In order to formulate rules (\ref{a1})--(\ref{a3}) below, we need to bound the
weights.
Thus, we choose $-1\leq w_{l, j}, c_j\leq 1$. 
Given are $m$ input vectors $\vec{x}_1,\dots,\vec{x}_m$ with $d$
nonnegative components each. We denote component $j$ of $\vec{x}_k$ with $
x_{k,j}\geq 0$. 
The weights have to be chosen such that the $m$ output vectors
$\vec{o}_1,\dots,\vec{o}_m$ are closest to given target vectors
$\vec{t}_1,\dots,\vec{t}_m$ in the $l^1$ norm $\sum_{k=1}^m \sum_{j=1}^n
|o_{k,j}-t_{k,j}|$.
To this end, we express difference $o_{k,j}-t_{k,j}$ via two nonnegative
variables $\delta_{k,j}^{+}, \delta_{k,j}^{-}\geq 0$:%\vspace*{-1em}
\begin{equation}
o_{k,j}-t_{k,j} = \delta_{k,j}^{+} - \delta_{k,j}^{-}.\label{a7}
\end{equation}
This leads to the problem%\vspace*{-1em}
\begin{equation}
\text{minimize} \sum_{k=1}^m \sum_{j=1}^n  (\delta_{k,j}^{+} +
\delta_{k,j}^{-})\label{minprob}
\end{equation}
under following restrictions (\ref{defa}), (\ref{a1}), (\ref{a3a}), and
(\ref{a3}) that deal with computing $o_{k,j}$.
For each $k\in[m]$ and $j\in[n]$ we compute $o_{k,j} = \sigma\left( a_{k,j} 
\right) \geq 0$,
\begin{alignat}{1}
a_{k,j} &:=c_j+\sum_{i=1}^d  w_{j,i} x_{k,i},\label{defa}
\end{alignat}
where
$\sigma(x)$
is the ReLU function.
Let $\tilde{M}:=\max\{ x_{k,j} : k\in[m],\, j\in [d]\}$. 
Both $o_{k,j}$ and $|a_{k,j}|$ are bounded by $d\tilde{M}+1$.
In Section \ref{secMILP2} we determine new inputs not
necessarily bounded by $\tilde{M}$ but by $1.1 \cdot \tilde{M} + 0.1$. Thus, values of $o_{k,j}$ and
$|a_{k,j}|$ are generally bounded by 
$$M:=d\cdot (1.1\cdot \tilde{M}+0.1)+1.$$

To implement the piecewise definition of ReLU, we
introduce binary variables $b_{k, j}$ that model, for input
$\vec{x}_k$, whether a neuron $j$ fires (value 1) or does not fire (value 0),
i.e., if the input of ReLU exceeds zero (cf.~\cite{Fischetti18},  \cite{Khalil18},
\cite{Serra17}):
\begin{equation}
-M(1-b_{k,j})  \leq a_{k,j} \leq M b_{k,j}.
\label{a1}
\end{equation}
If $b_{k,j}=1$, output $o_{k,j}$, $0\leq
o_{k,j}\leq M$, of neuron $j$ equals $a_{k,j}$ for input $\vec{x}_k$. Otherwise
for $b_{k,j}=0$, the output $o_{k,j}$ has to be set to zero:
\begin{alignat}{1}
 - M (1-b_{k,j}) &\leq o_{k,j}-a_{k,j}
 \leq M(1-b_{k,j}),\label{a3a}\\
 0 &\leq o_{k,j} \leq M b_{k,j}.\label{a3}
\end{alignat}

%-b_{k,j}M \leq o_{k,j} \leq b_{k,j}M.\label{a3}

%To avoid a mixed integer program, one could split up
%\begin{equation}
%a_{k,j} = o_{k,j}-a_{k,j}^{-}\label{split}
%\end{equation}
%with nonnegative variables $a_{k,j}^{-}$. 
%To obtain the correct value of $o_{k,j}$ one has to minimize $a_{k,j}^{-}$
%s.t.~(\ref{split}) for a given value of $a_{k,j}$ (which unfortunately is not
%fixed).
%
%Thus, one could use objective
%function
%\begin{equation*}
%\text{minimize} \sum_{k=1}^m \sum_{j=1}^n  \delta_{k,j}^{+} +
%\delta_{k,j}^{-} + a_{k,j}^{-}.
%\end{equation*}
%However, minimization of $ \delta_{k,j}^{+} +
%\delta_{k,j}^{-}$ and of $a_{k,j}^{-}$ might influence each other.

The MILP can be divided into $n$ independent MILPs that calculate
$d+1$ weights separately for each of the $n$ neurons of the layer.

In theory, these MILPs can be replaced by $2^m$ LPs as follows. Let $j\in[n]$.
For each $k\in [m]$, we can add
constraints $a_{k,j} < 0$ or $a_{k,j}\geq 0$ to
avoid binary variables and obtain LPs. Then, the weights are determined by a
smallest objective value of all problems.

We really replace the MILP of the last layer by
a single LP, which is potentially much faster than the MILP:
To test the network, we use ground truth data consisting of one-hot vectors.
If the digit to be recognized is $j$, $0\leq j\leq 9$, then the $j$th component
is one, all other components are zero. 
Since the prediction is an output closest to one, we can replace ReLU with the
identity function to obtain a linear problem, i.e.
$o_{k,j}:=a_{k,j}$ without constraints (\ref{a1})--(\ref{a3}) such that
$o_{k,j}$ may be negative. Instead of objective function
(\ref{minprob}) we deal with
\begin{equation}
\text{minimize} \sum_{k=1}^m \sum_{j=1}^n  \delta_{k,j},\label{targetlast}
\end{equation}
where
\begin{equation}
\delta_{k,j} := \left\{\begin{array}{ccc}
 \delta_{k,j}^{+}  &:& \text{ ground truth } t_{k,j}=0\\
\delta_{k,j}^{+} +
\delta_{k,j}^{-} &:& \text{ ground truth }  t_{k,j}=1.
\end{array}
\right.\label{targetlast2}
\end{equation}
In the linearized version (\ref{targetlast}) of the error functional, we do not
consider $\delta_{k,j}^{-}$ in the case $t_{k,j}=0$ because only positive ReLU inputs
$a_{k,j}$ contribute to the error. Non-positive inputs would be set to
zero by applying the ReLU function and then match $t_{k,j}=0$. 

\subsection{Proposing layer inputs}\label{secMILP2}
After optimizing the weights of a layer, its input data are slightly
adjusted to further minimize the output
error of that layer. This is described in what follows.

To adjust the input of a layer, basically the same MILP/LP as before can be
used.
Now weights $W\in\mathbb{R}^{n\times d}$ and
$\vec{c}\in\mathbb{R}^{n}$ of one building block (see Figure
\ref{figureblock}) with $d$ inputs and $n$ outputs are given and are not
variable.
We need to find $m$ input vectors
$\vec{x}_1,\dots,\vec{x}_m$ each with $d$ components (which are now variables $x_{k,j}\geq 0$, 
$k\in [m]$, $j\in [d]$), so that for given weights, the problem
(\ref{minprob}) is solved under constraints (\ref{a7}), (\ref{defa}),
(\ref{a1}),  (\ref{a3a}), and (\ref{a3}) for all but the last layer. For the
last layer, problem (\ref{targetlast}) is solved under restrictions
$o_{k,j}=a_{k,j}$, (\ref{a7}), (\ref{defa}), (\ref{targetlast2}).
Only small adjustments of inputs promise not to lead
to major changes in the weights in subsequent steps. This is important since
we do not want to forget information that has already
been learned.
Let $\tilde{x}_{k,j}$ be the input of the layer previous to this optimization
step.
Then we add bounds
\begin{equation}
\max\{0, 0.9\cdot \tilde{x}_{k, j}-0.1\} \leq x_{k, j} \leq
1.1\cdot \tilde{x}_{k, j}+0.1.
\label{boundsweights}
\end{equation}
The bounds are helpful beyond that. Because without them, the runtime for
determining subsequent weights increases significantly.
Inputs can be calculated independently for each of the $m$ training input
vectors. 

Instead of adjusting inputs to optimally match desired
outputs of one single layer, an alternative approach would be to
consider all subsequent layers with the goal of minimizing the distance to the ground truth. With weights held
fixed, this is a linear problem similar to the tasks in
\cite{Fischetti18,Khalil18}, etc.
However, it turned out that considering more than one layer is not
necessary due to the chosen iterative approach.
\subsection{Post-processing of the weights of the last
layer}\label{seclastlayer} So far, the objective functions have been built on the $l^1$
norm, which is needed in particular for weight calculation of
hidden layers.
But now, in a final step (see Algorithm \ref{algMILP}), we adjust the weights of
the last layer with an LP by minimizing $ \sum_{k=1}^m \sum_{j=1}^n  (\delta_{k,j} - s_{k,j})
$
%todo: Auch hier negative Variable fuer Bound nicht verwenden
where variables $\delta_{k,j}$ are defined in (\ref{targetlast2}), and
slack variables $0\leq  s_{k,j} \leq 0.49$ are additionally constrained by
$  s_{k,j} \leq \delta_{k,j}$.
Thus, we allow deviations 
%from ground truth values 
up to $0.49$
so that zeroes and ones of ground truth vectors are still separated.
%
% such that output components in $[0,0.5)$ represent
%zero and values in $(0.5, 1.5)$ represent one.
%
%If one uses this objective function to determine weights of the last layer
%within iterations, runtime increases heavily.
%
\section{Results and batch learning}
\begin{figure*}[htb]
\begin{center}
     {\setlength{\epsfxsize}{1.0\columnwidth}\epsfbox{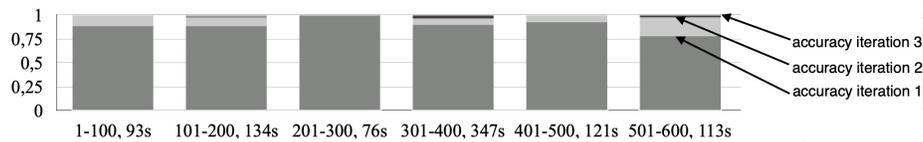}}
 \end{center}    
\caption{For training with 100 referenced MNIST images and randomly
initialized weights, vertical bar segments show how (while-) iterations in
Algorithm \ref{algMILP} increase the accuracy of the 784-8-8-8-10
network until a final accuracy of one is reached. The
accuracy after the first iteration is shown at the bottom. Then the improvement
of each subsequent iteration is added.
The post-processing step was not required. Runtimes were 
measured with CPLEX 12.8.0 on a MacBook Pro with 16 GB RAM and an i5 processor
(two cores).
\label{figuretraining0}}
\end{figure*}
%\begin{figure}[htb]
%\begin{center}
%{\setlength{\epsfxsize}{0.99\columnwidth}\epsfbox{./training1.eps}}
% \end{center}     
%\caption{Iterative training of 15 batches with 100
%consecutive MNIST images each, i.e. with the first 1500 images of
%the training dataset: The top curve shows training
%accuracy with respect to each single batch. The middle curve
%represents the accuracy with respect to all training data so far seen. This
%consists of the current and all preceding batches. The lower curve visualizes
%the accuracy on the MNIST test dataset with 10,000 images.
%Runtime based on CPLEX 12.8.0 was 17,455 s on a Macbook Pro with i5
%processor (two cores) and 16 MB RAM. A timelimit of 600 s processor time was
% exceeded in ten MILPs.
%\label{figuretraining1}}
%\end{figure}

\begin{figure*}[htb]
\begin{center}
      {\setlength{\epsfxsize}{1.0\columnwidth}\epsfbox{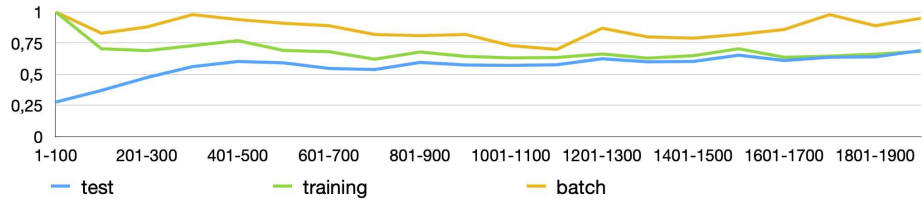}}
\end{center}    
\caption{Iterative training of the 784-8-8-8-10 network with 20 batches of 100
consecutive images. Weights are bounded due to (\ref{wbound}).
The top curve shows the training
accuracy with respect to each single batch. The middle curve
represents the accuracy with respect to all training data seen so far. This
consists of the current batch and all previous batches. The bottom curve
visualizes the accuracy on the MNIST test dataset with 10,000 images. The
runtime was 1,448 s.
\label{figuretraining2}}
\end{figure*}

Due to runtimes of MILPs, we did not
apply all steps of Algorithm \ref{algMILP} to all 60,000 training images but
only to small subsets (batches) of one hundred images. However, the LP of
the post-processing step is able to handle the complete training
set.
The outcome of Algorithm \ref{algMILP} depends strongly on the initialization of
weights. 
%
%For example, if one initially sets all weights zo zero, then the input
%of the last layer is a zero vector and optimization can't yield useful results. 
%By randomly initializing weights
%$w_{j, i} \in [0, 0.1]$, $c_j\in [-0.05, 0.05]$ of the 784-8-8-8-10 network,
%Algorithm \ref{algMILP} on the first hundred MNIST training images performed
%four while-iterations (with intermediate accuracies of 0.33,
%0.58, 0.95) to reach an accuracy of
%one in 460 seconds processor time\footnote{CPLEX 12.8.0 on a Macbook Pro with
% i5 processor (two cores) and 16 MB RAM} without the need of the
%post-procession step. 
%
A good random initialization of
weights $w_{j, i}, c_j \in [-1, 1]$ leads to the results shown in 
Figure \ref{figuretraining0} for the 784-8-8-8-10 network.
While one can experimentally determine a
suitable initialization, a larger issue is that accuracy is low on
all 60,000 images after training on 100 images.
%(10.46\% for previous example). 
Therefore, we experimented with iterative batch learning.
%Figure \ref{figuretraining1} shows the outcome of iterative training with
%batches of 100 images. 
Algorithm \ref{algMILP} was applied to an initial batch of images 1-100, and
weights were updated accordingly. Then the algorithm was applied to
images 101-200 on these updated weights, etc.
We use a simple idea to better remember previously learned images: We do not
only initialize weight variables for a warm
start with values from the preceding batch training, but we also
limit weight changes of consecutive
batch learning steps. 
%so that weights might still work well for previous batches.
Starting with the training of the second batch,
each weight $w:=w_{j,i}$ or
$w:=c_j$ is additionally bounded depending on the corresponding computed weight
$\tilde{w}$ of the same layer for the preceding batch (factor 0.6 was
determined experimentally):
\begin{alignat}{1}
 \tilde{w}-0.6\cdot|\tilde{w}|-0.01 &\leq
 w \leq
 \tilde{w}+0.6\cdot|\tilde{w}|+0.01.\label{wbound} 
\end{alignat}
%Constants were determined experimentally. 
%Factor 0.6 appears to be quite
%large, but experiments showed that smaller values prohibited adjustment to new
%batch data.
%------------------
This approach yielded a best case
accuracy of $0.69$ on the 10,000 test images 
(while Tensorflow/Keras reached a maximum of 73.3\% after three epochs
with the Adam optimizer, %\cite{Adam}, 
learning rate 0.001, batch size 100, all random
seeds set to 4711),
%by using softmax activation on the last layer, this increases to 93,24\% after
% 37 epochs),
%GBB Ende
see Figure \ref{figuretraining2}.
Bound (\ref{wbound})
also reduced processing times.
\begin{figure}[tb]
%\vspace*{-1em}
\begin{center}
\begin{tabular}{|c|c|c|c|c|c|}\hline
             \setlength{\epsfxsize}{0.11\columnwidth}\epsfbox{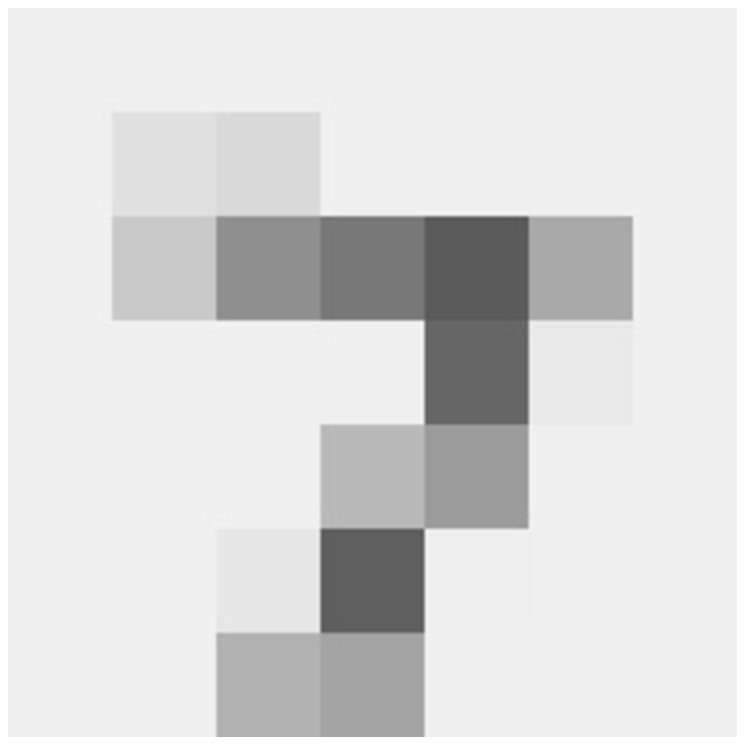}&
             \setlength{\epsfxsize}{0.11\columnwidth}\epsfbox{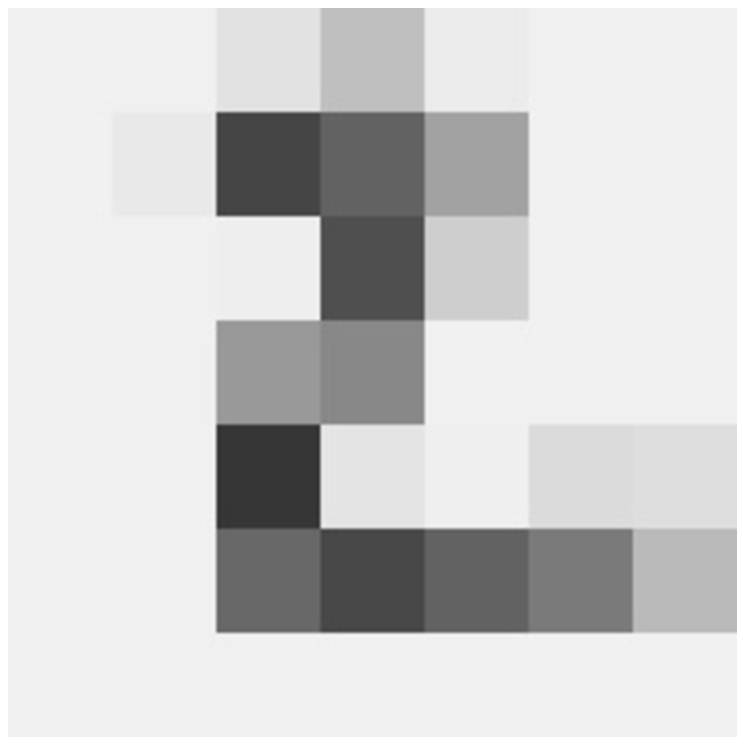}
             &
             \setlength{\epsfxsize}{0.11\columnwidth}\epsfbox{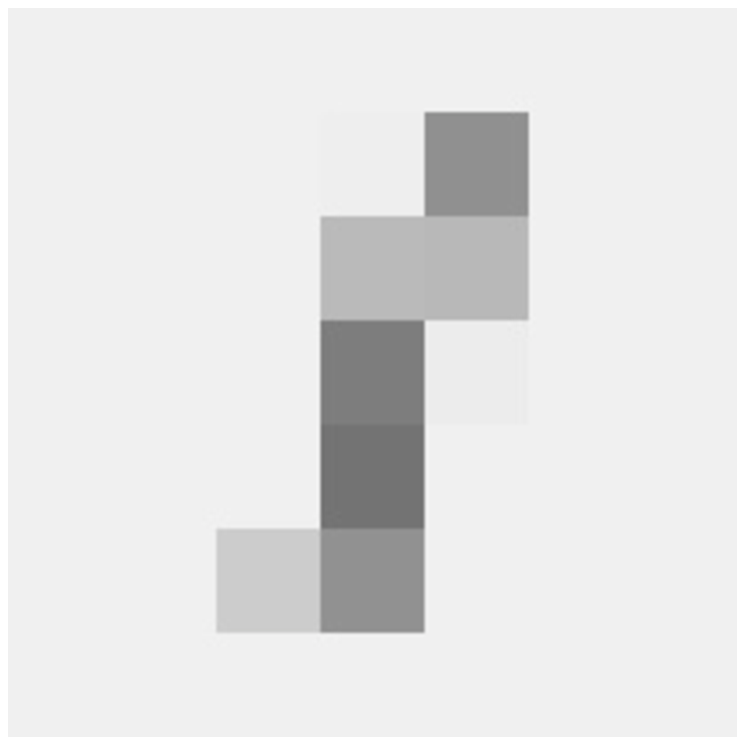}& 
             \setlength{\epsfxsize}{0.11\columnwidth}\epsfbox{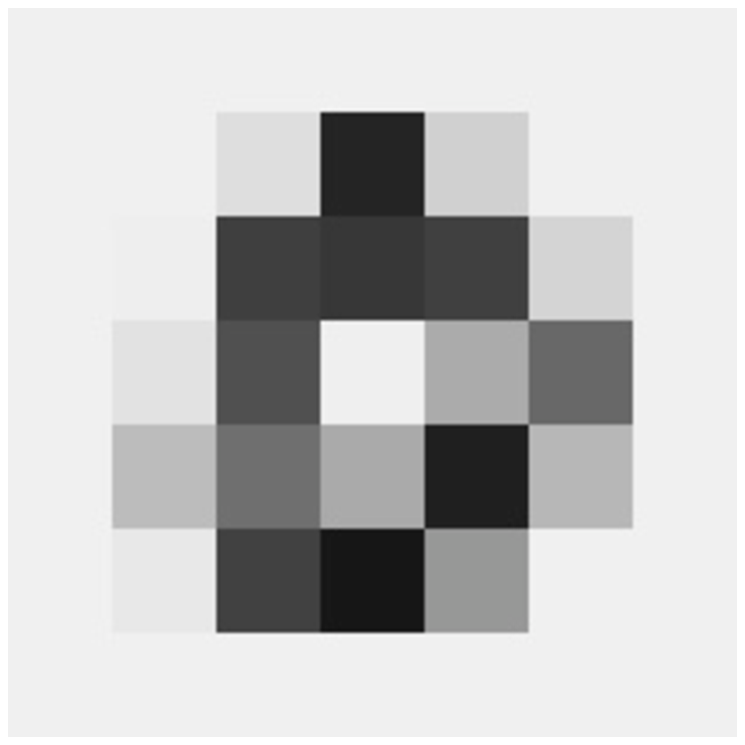}
             &
             \setlength{\epsfxsize}{0.11\columnwidth}\epsfbox{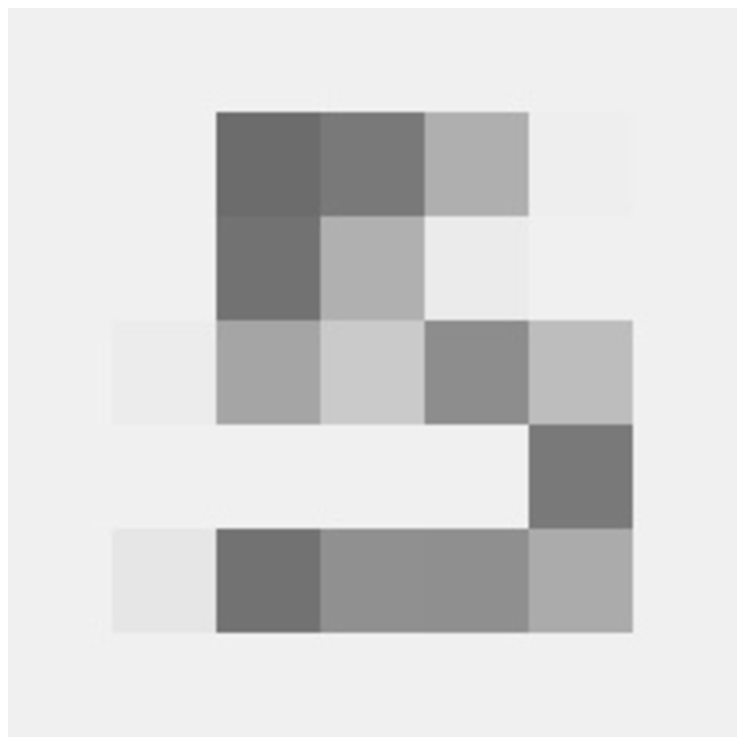}
             &
             \setlength{\epsfxsize}{0.11\columnwidth}\epsfbox{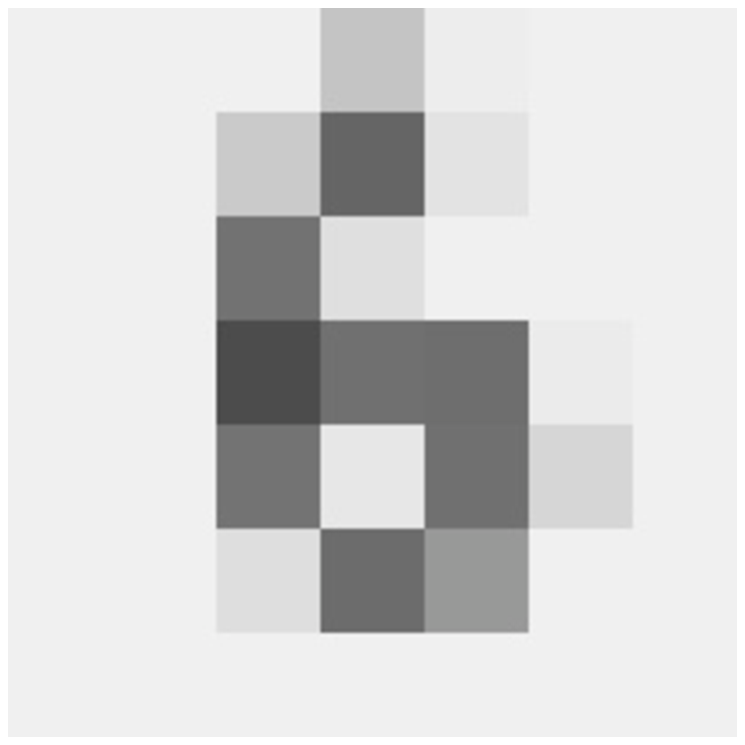}
             \\
			 7 (7) & 2 (2) & 1 (1) & 0 (0) & 5 (5) & 6 (6)\\
			 \hline
\end{tabular}	
\quad
\begin{tabular}{|c|c|c|c|c|c|}\hline
			 \setlength{\epsfxsize}{0.11\columnwidth}\epsfbox{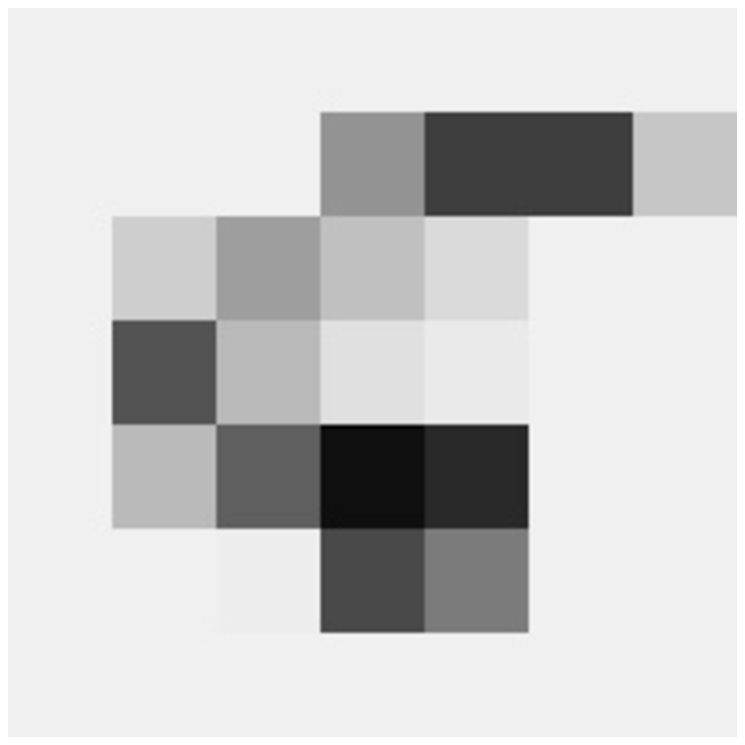} &
			 \setlength{\epsfxsize}{0.11\columnwidth}\epsfbox{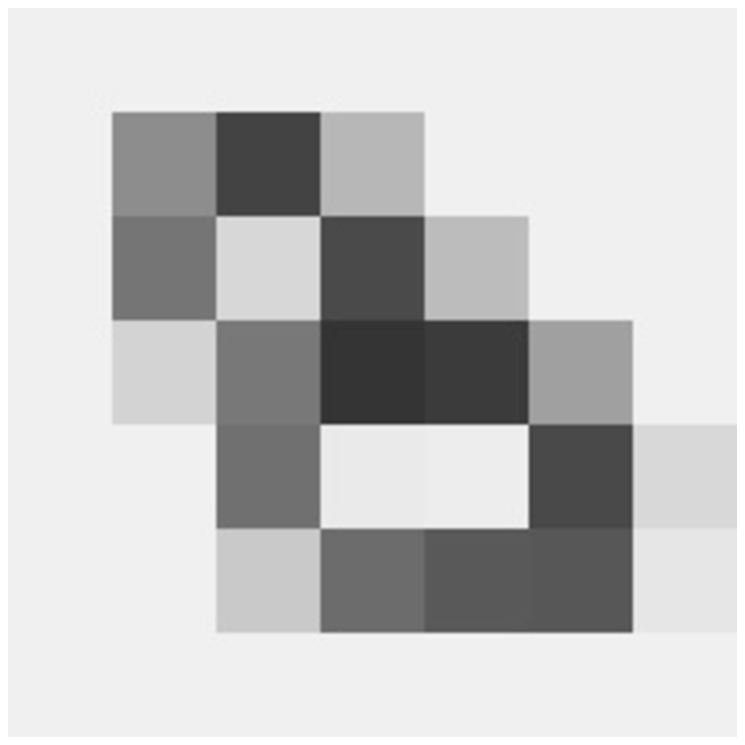} &
			 \setlength{\epsfxsize}{0.11\columnwidth}\epsfbox{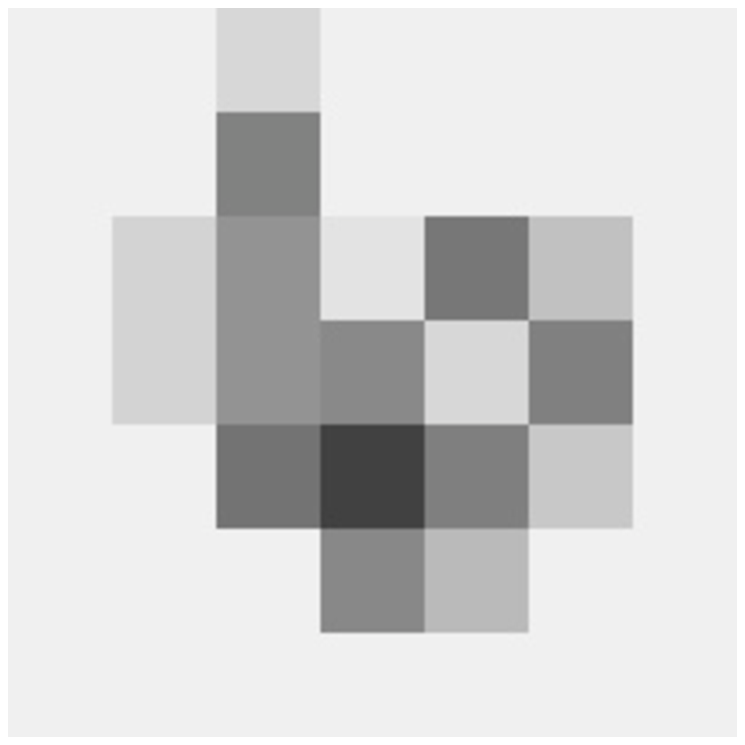} &
			 \setlength{\epsfxsize}{0.11\columnwidth}\epsfbox{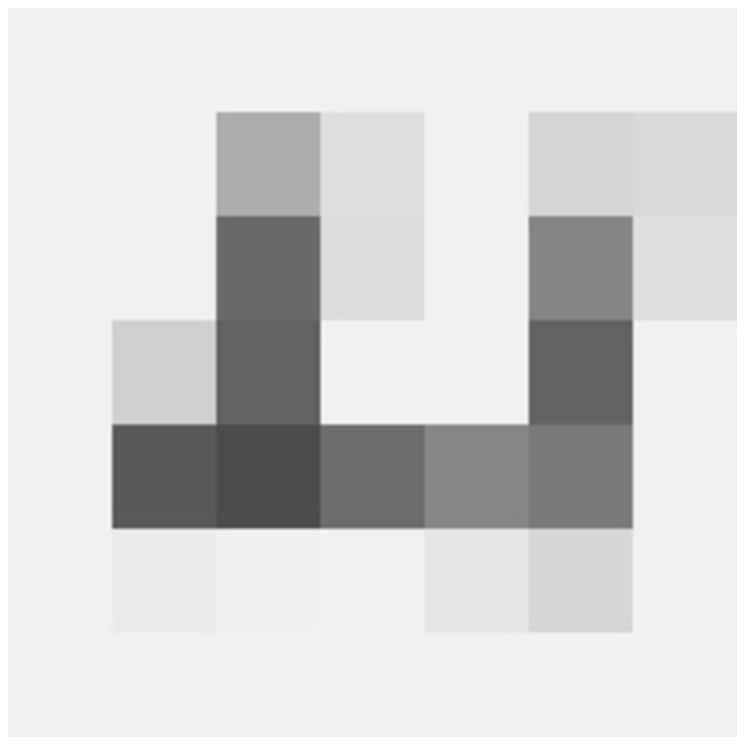} &
			 \setlength{\epsfxsize}{0.11\columnwidth}\epsfbox{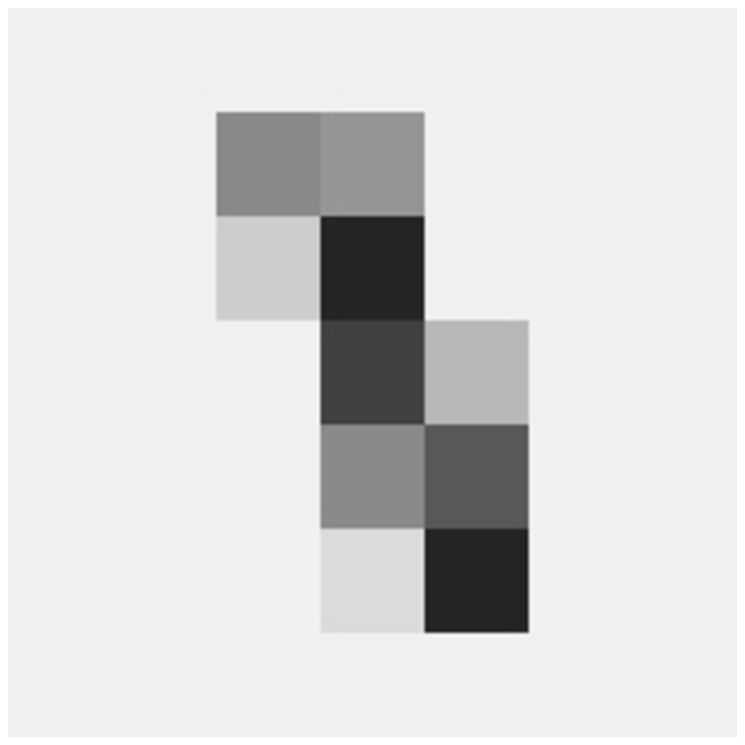} &
			 \setlength{\epsfxsize}{0.11\columnwidth}\epsfbox{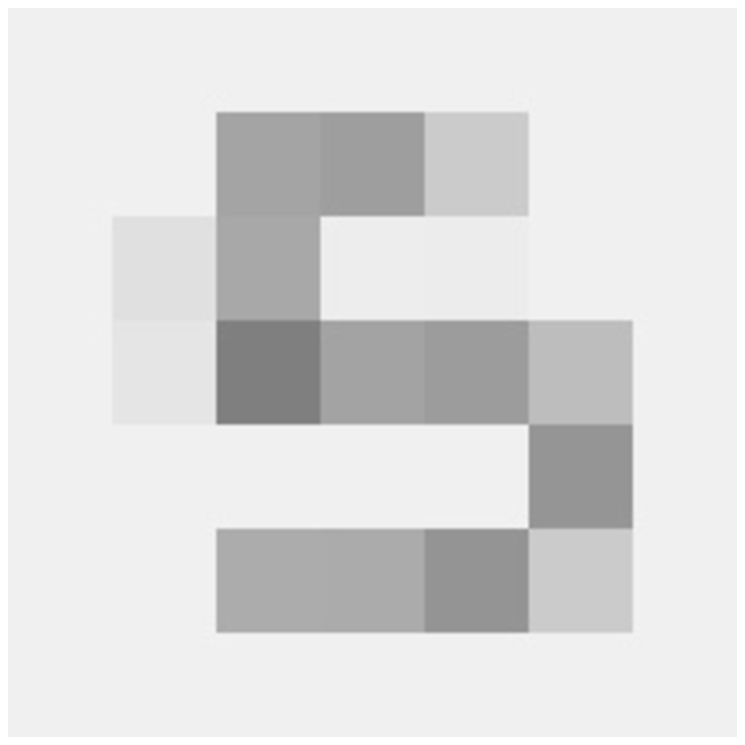} \\
			 4 (5) & 2 (3) & 4 (6) & 6 (4) & 3 (1) & 6 (5)\\
			 \hline
\end{tabular}	
\end{center}
\caption{Reducing the resolution contributes to false
detections. Ground truth is given in brackets. The best detection results (IoU
values) were obtained for digit 1, worst for 5.
\label{figurewrong}}
\end{figure}
%
%\begin{figure}[htb]
%\begin{center}
%      {\setlength{\epsfxsize}{0.99\columnwidth}\epsfbox{./training3.eps}}
% \end{center}     
%\caption{Bound (\ref{wbound}) reduces processing times and
%allows for larger batches. The first batch contains 100 images.
%Starting with the second batch, 200 images are processed. In total, 1900 images
%were considered for learning. Meaning of curves is as
%in Figure \ref{figuretraining1}. Runtime was
%8,678 s.
%\label{figuretraining3}}
%\end{figure}
%
%So far, discussed results were obtained with only one epoch of training. 
We also
trained with multiple epochs on shuffled data. To avoid overfitting, we
added noise to ground truth vectors and tested a dropout strategy. However, all
these methods did not significantly improve accuracy - unlike running
the LP of the post-processing step on all 60,000 training images. To this end,
we did not apply it for
each batch but ran it after completing ten batches (i.e.,
training on 1000 images).
It then achieved test accuracies of up to 75.84\% within about a thousand
seconds processor time.
A majority vote of a committee of three
networks trained on different sets of 1,000 images increased the accuracy to
79.19\%. 
%
%
%----------------------------------
%todo: entfernen, falls mehr als 6 Seiten

For the convolutional network (randomly initialized with weights in $[0, 1]$), we similarly 
trained weights iteratively on ten batches of 100 images (first 1000 images of training set) with rule
(\ref{wbound}) and then ran the post-processing step (Algorithm \ref{algMILP}, 
Section \ref{seclastlayer}) on all 60,000 training images in 3,933
seconds processor time to achieve an accuracy of 87.11\%.
%
%
%Thus, accuracy can be increased only slightly by taking a
%majority vote of multiple networks.
Downsampling of image resolution
was necessary to run MILPS in reasonable time, but reduces accuracy, 
see Figure \ref{figurewrong}.
Using softmax activation on the last layer, the Adam optimizer
(with parameters as before)
achieved an accuracy of 93,51\% on test data within 40 epochs 
when trained with full-resolution $28\times 28$ images. However, after adding an
average pooling layer to reduce resolution consistent with MILP training to
$7\times 7$ pixels, only 89,61\% accuracy is obtained in one minute (40 epochs).
ReLU instead of softmax activation on the last layer implied worse accuracies up
to 46\%, thus MILP training performed better.

In this report, we have shown that it is possible to train networks
iteratively based on MILPs. 
Accuracies as with the Adam Optimizer can be achieved. Thus, combinatorial optimization could be an
alternative to gradient-based methods when they encounter difficulties.
However, without further consideration, runtimes currently limit this approach to small training 
sets and simple networks. 
Future work may be concerned with the improvement of runtimes.

\section*{Acknowledgements}
Many thanks to Christoph Dalitz for his valuable comments.
\bibliographystyle{spmpsci}
\bibliography{neural}
\end{document}